%% file: main.tex
\documentclass[runningheads]{llncs}
\usepackage{amsmath}

\usepackage[T1]{fontenc}
\input{sections/macros.tex}

\begin{document}
	
\renewcommand{\paragraph}[1]{\medskip\noindent\textbf{\emph{#1.}}}
\title{Efficient Algorithms for Partial Constraint Satisfaction Problems over Control-flow Graphs} 
\titlerunning{Partial Constraint Satisfaction Problems over CFG}
\author{Xuran Cai \and Amir Goharshady}
\authorrunning{X.~Cai \and A.K.~Goharshady}
%
\institute{University of Oxford\\
Oxford, United Kingdom\\
\email{xuran.cai@cs.ox.ac.uk}, \email{amir.goharshady@cs.ox.ac.uk}}

\maketitle              

\input{sections/abstract}
\input{sections/introduction}

\input{sections/SPLDecom}
\input{sections/PCSP}

\input{sections/bankSelection}

\input{sections/experiments}

\input{sections/conclusion}

\bibliographystyle{splncs04}
\bibliography{sections/refs}

\end{document}

%% file: sections/macros.tex
\usepackage{rotating}
\usepackage{amsthm}
\usepackage{amsfonts}
\usepackage{nicefrac}
\usepackage{wrapfig}
\usepackage{amscd}
\usepackage{verbatim}
\usepackage{tikz}
\usepackage{paralist}
\usepackage{microtype}
\usepackage{makecell}
\usepackage{listings}
\usepackage{caption}
\usepackage{xcolor}
\usepackage{color,soul}
\usepackage{amssymb}

\usetikzlibrary{positioning}
\usepackage{adjustbox}
\usepackage{algpseudocode}
\usepackage[ruled,linesnumbered]{algorithm2e}
\usepackage{setspace}
\usepackage{pgf}
\usepackage{graphicx}
\usepackage{dashbox}
\usepackage{float}
\usepackage{syntax}
\usepackage{hyperref}
\usepackage{hhline, booktabs}
\usepackage{array,multirow}	
\usepackage{tablefootnote}

\usepackage{amsmath}
\usepackage{thmtools}

\usetikzlibrary{arrows}
\usetikzlibrary{positioning}
\usepackage{ textcomp }
\usepackage{subcaption}
\tikzset{every node/.append style={transform shape}}

\definecolor{codegreen}{rgb}{0,0.6,0}
\definecolor{codegray}{rgb}{0.5,0.5,0.5}
\definecolor{codepurple}{rgb}{0.58,0,0.82}
\definecolor{backcolour}{rgb}{0.95,0.95,0.92}

\renewcommand\thmcontinues[1]{Continued}

\newcommand{\sloop}{\circledast}

\newcommand{\oseries}[2]{#1 \otimes #2}
\newcommand{\oparal}[2]{#1 \oplus  #2}
\newcommand{\oloop}[1]{#1^\circledast}

\newcommand{\cfg}[1]{\textsf{cfg}(#1)}
\newcommand{\lt}[1]{\textsf{lt}(#1)}

\newcommand{\vars}{\mathbb{V}}
\newcommand{\cost}{\textsc{Cost}}
\newcommand{\DP}{\texttt{dp}}
\newcommand{\compatible}{\leftrightharpoons}
\usepackage{amsmath}

\tikzset{
    arrow/.style={thick,->,>=stealth, draw=blue!20!black},
    vertex/.style={rectangle, minimum width=10mm, minimum height=10mm, fill=green!10, draw=black!50},
    whitevertex/.style={rectangle, minimum width=10mm, minimum height=10mm, fill=yellow!10, draw=black!50},
    grayvertex/.style={rectangle, minimum width=10mm, minimum height=10mm, fill=gray!20, draw=gray!80},
    redvertex/.style={rectangle, minimum width=10mm, minimum height=10mm, fill=orange!20, draw=orange!80},
    bluevertex/.style={rectangle, minimum width=10mm, minimum height=10mm, fill=green!30, draw=green!80},
	vertexs/.style={thick, rectangle, minimum size=10mm, fill=green!20, draw=black},
    vertext/.style={thick, rectangle, minimum size=10mm, fill=red!20, draw=black},
    vertexc/.style={thick, rectangle, minimum size=10mm, fill=green!10, draw=black},
    vertexb/.style={thick, rectangle, minimum size=10mm, fill=blue!20, draw=black, text=black},
    vertexla/.style={minimum size=5mm, draw=none, text=red},
    emptynode/.style={thick, rectangle, minimum size=10mm, fill=white, draw=white, text=white},
    rect/.style={thick, rectangle, minimum size=10mm, fill=black!5, text=black}
}

%% file: sections/abstract.tex
\begin{abstract} 
   
   In this work, we focus on the Partial Constraint Satisfaction Problem (PCSP) over control-flow graphs (CFGs) of programs.  PCSP serves as a generalization of the well-known Constraint Satisfaction Problem (CSP). In the CSP framework, we define a set of variables, a set of constraints, and a finite domain $D$ that encompasses all possible values for each variable. The objective is to assign a value to each variable in such a way that all constraints are satisfied. In the graph variant of CSP, an underlying graph is considered and we have one variable corresponding to each vertex of the graph and one or several constraints corresponding to each edge. In PCSPs,  we allow for certain constraints to be violated at a specified cost, aiming to find a solution that minimizes the total cost.  Numerous classical compiler optimization tasks can be framed as PCSPs over control-flow graphs. Examples include Register Allocation, Lifetime-optimal Speculative Partial Redundancy Elimination (LOSPRE), and Optimal Placement of Bank Selection Instructions. On the other hand, it is well-known that control-flow graphs of structured programs are sparse and decomposable in a variety of ways. In this work, we rely on the Series-Parallel-Loop (SPL) decompositions as introduced by~\cite{RegisterAllocation}. Our main contribution is a general algorithm for PCSPs over SPL graphs with a time complexity of \(O(|G| \cdot |D|^6)\), where \(|G|\) represents the size of the control-flow graph. Note that for any fixed domain $D,$ this yields a linear-time solution. Our algorithm can be seen as a generalization and unification of previous SPL-based approaches for register allocation and LOSPRE. In addition, we provide experimental results over another classical PCSP task, i.e. Optimal Bank Selection, achieving runtimes four times better than the previous state of the art.

       \keywords{Structured Programs \and Compiler Optimization \and Control-flow Graphs \and Graph Decompositions\and Partial Constraint Satisfaction Problems.}

\end{abstract}

%% file: sections/introduction.tex
\section{Introduction}

\paragraph{CSP and PCSP}
Constraint Satisfaction Problems (CSPs) provide an expressive framework for a wide variety of tasks in different fields, especially in compiler optimization. CSPs involve determining values for variables while adhering to constraints that define permissible combinations of these values~\cite{PCS}. Many common graph-related problems, such as the graph coloring problem, can be formulated as CSPs. However, there are instances where it may be infeasible or impractical to find complete solutions that satisfy every constraint. In such cases, we may aim for partial solutions, specifically by satisfying the maximum number of constraints or assigning a cost to each unsatisfied constraint and minimizing the total cost, which comes to the area of Partial Constraint
Satisfaction Problems(PSCPs)~\cite{Koster2002SolvingCS}.

PCSPs have a wide array of applications. A notable example that translates elegantly into the PCSP framework is the MAX-SAT problem~\cite{KOSTER199889}. Additionally, various compiler optimization tasks, particularly those related to graph theory, can be represented as PCSPs. Examples include register allocation~\cite{RegisterAllocation}, lifetime-optimal speculative partial redundancy elimination~\cite{cai2024faster}, and optimal placement of bank selection instructions~\cite{bankselection}.

The NP-hardness of PCSPs, even with a domain size of $3,$ is established through an easy reduction from the 3-coloring problem in graphs. In~\cite{KOSTER199889}, it was shown via a reduction from MAX-SAT that PCSPs are NP-hard even when all domains are restricted to size two. Computational experiments support these findings in practical scenarios~\cite{KOSTER199889}.

\paragraph{Efficient PCSP Algorithms}
Although both CSP and PCSP are NP-hard, efficient polynomial-time algorithms exist when the underlying primal graph is a tree or has a tree-like structure~\cite{DECHTER19871,freuder1990complexity,MACKWORTH198565}. Even in the absence of such structures, recognizing that a problem can be viewed as containing a tree or being part of a tree-like structure remains beneficial and leads to effective heuristics~\cite{DECHTER19871,10.5555/2887965.2887992}. A significant advancement in PCSP algorithms was obtained by~\cite{Koster2002SolvingCS}, utilizing tree decompositions. This approach provides an efficient parameterized algorithm based on the treewidth of the graph. Treewidth is a measure of tree-likeness. Intuitively speaking, graphs of treewidth $t$ can be decomposed into small sets of vertices, each of size at most $t+1$ that are connected to each other in a tree-like formation. 

\paragraph{Control-flow Graphs and their Sparsity}
The control-flow graph (CFG) of a program is defined as a graph where each vertex represents a statement in the program, and a directed edge exists between two vertices if their corresponding statements can be executed in succession. Some applications adopt a slightly coarser definition of CFGs, where each vertex corresponds to a basic block of program statements. In both instances, it is well established that the control-flow graphs of real-world programs are sparse, often resembling trees, and can be decomposed into sets of vertices, each containing at most 8 vertices that are interconnected in a tree-like fashion. More formally, CFGs of structured goto-free programs have a treewidth of at most 7~\cite{treewidth}. This notable result has been leveraged in various areas, including program analysis, compiler optimization, and model checking, where the small treewidth property enables faster algorithms for $\mu$-calculus model checking~\cite{DBLP:conf/cav/Obdrzalek03}, data-flow analysis~\cite{DBLP:conf/esop/ChatterjeeGIP20,DBLP:conf/vmcai/GoharshadyZ23}, Markov Decision Processes (MDPs)~\cite{DBLP:conf/cav/ChatterjeeL13,DBLP:conf/atva/AsadiCGMP20,DBLP:conf/fsttcs/AhmadiCGMSZ22}, reachability analysis~\cite{DBLP:conf/cav/000120,DBLP:journals/toplas/ChatterjeeGGIP19}, algebraic program analysis~\cite{DBLP:journals/pacmpl/ConradoGKTZ23,DBLP:journals/toplas/ChatterjeeIGP18,DBLP:conf/popl/ChatterjeeGIP16}, register allocation~\cite{DBLP:conf/cc/Krause13,DBLP:conf/soda/BodlaenderGT98}, cache management~\cite{DBLP:conf/pldi/AhmadiDGP22,DBLP:journals/pacmpl/ChatterjeeGOP19}, and equality saturation~\cite{oopsla24}.

The original treewidth bound established in~\cite{treewidth} applied to Pascal and C programs, but subsequent research has extended this result to other programming languages, including Ada~\cite{DBLP:conf/adaEurope/BurgstallerBS04}, Java~\cite{DBLP:conf/alenex/GustedtMT02,DBLP:conf/atva/ChatterjeeGP17}, and Solidity ~\cite{DBLP:conf/sac/ChatterjeeGG19}, as well as to path decompositions~\cite{DBLP:journals/pacmpl/ConradoGL23}. However, there are also negative findings indicating that bounded treewidth does not always facilitate verification~\cite{DBLP:conf/lpar/FerraraPV05}.

In the context of register allocation, the recent work~\cite{RegisterAllocation} introduced a new decomposition concept known as Series-Parallel-Loop (SPL), which precisely captures the set of control-flow graphs for structured goto-free programs and formalizes their sparsity. We build upon the same decomposition concept but show that it has a much higher potential than the authors of~\cite{RegisterAllocation} imagined and can be exploited to solve a significantly wider family of problems than the register allocation task investigated in~\cite{RegisterAllocation}.

\paragraph{Our Contribution}
In this work, we focus on the general family of Partial Constraint Satisfaction Problems (PCSPs) over the CFGs of structured goto-free programs. Our approach is a generalization of~\cite{RegisterAllocation} and~\cite{cai2024faster}, and thus contains register allocation and LOSPRE as special cases. We develop a \textit{linear-time} algorithm for PCSPs. Specifically, we leverage the sparsity of CFGs and their SPL decompositions to design an algorithm with a runtime of  \( O(n \cdot |D|^6) \), where $n$ is the number of lines in the program and $D$ is the domain of variables. Unlike the method in \cite{Koster2002SolvingCS}, we do not rely on tree decompositions. Thus, there is no extra parameter such as treewidth. Instead, we use SPL decompositions~\cite{RegisterAllocation}. This choice results in a significantly simpler algorithm since SPL decompositions accurately represent the set of CFGs, allowing us to avoid solving the problem on a broader set of graphs than necessary. Moreover, SPL decompositions take into account the direction of edges in the CFG.

On the experimental side, we apply our general solution to the problem of Optimal Placement of Bank Selection Instructions, comparing it with the state-of-the-art approach of~\cite{bankselection}. The simplicity of our algorithm yields practical benefits. We present extensive experimental results conducted on the Small Device C Compiler (SDCC), a highly optimized compiler based on~\cite{sdcc1,sdcc2}. Our algorithm achieves substantial performance improvements over~\cite{bankselection}.

\paragraph{Organization} Section~\ref{sec:spl} presents a formal definition of SPL decompositions, following~\cite{RegisterAllocation}. Section~\ref{sec:pcsp} covers our main contribution: a linear-time algorithm for general PCSPs over control-flow graphs that exploits SPL decompositions. This is followed by the example use-case in Bank Selection in Section~\ref{sec:bank}. Finally, Section~\ref{sec:exp} reports our experimental results.

%% file: sections/SPLDecom.tex
\section{SPL Decompositions} \label{sec:spl}

We will build our algorithm on top of a decomposition method introduced in~\cite{RegisterAllocation}. This decomposition method is called SPL (Series-Parallel-Loop)
and is an extension of series-parallel graphs with an extra loop operation. It is shown in~\cite{RegisterAllocation} that a graph is a CFG of a structured program if and only if it has an SPL decomposition. 

\paragraph{Structured Programs~\cite{treewidth}} We say a program is structured if it can be generated using the following grammar:
    \begin{equation} \label{gram:prog}
        \begin{array}{rl}
    P := & \epsilon \mid \texttt{break} \mid \texttt{continue}  \mid P ; P\\ & \mid \texttt{if} ~\varphi~ \texttt{then} ~P~ \texttt{else} ~P~ \texttt{fi} \mid \texttt{while} ~\varphi~ \texttt{do} ~P~ \texttt{od}.
    \end{array}
    \end{equation}
Here, $\epsilon$ is any atomic operation that has no effect on control flow, such as an assignment to a variable. It is easy to define other structures such as \texttt{for} and \texttt{switch} as syntactic sugar. See~\cite{treewidth} for details. We say a program generated by the grammar above is \emph{closed} if every \texttt{break} and \texttt{continue} statement appears inside a \texttt{while} loop's body.

\paragraph{SPL Graphs~\cite{RegisterAllocation}} An SPL graph $G = (V, E, S, T, B, C)$ is a directed graph $(V, E)$ with four distinct special nodes $S, T, B, C \in V,$ which are respectively called the \emph{start}, \emph{terminate}, \emph{break} and \emph{continue} nodes, generated by the grammar below:
\begin{equation} \label{gram:spl}
	\begin{array}{rl}
		G := & A_\epsilon ~|~ A_{\texttt{break}} ~|~ A_{\texttt{continue}} ~|~ \oseries{G}{G} ~|~ \oparal{G}{G} ~|~ \oloop{G}
	\end{array}
\end{equation}
We now explain the atomic graphs and operations in this grammar.

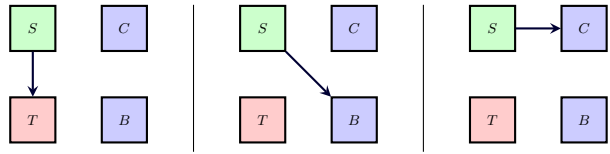
\begin{figure}[H]
	\begin{center}
	\begin{tabular}{cc @{\hspace{0.5cm}} |@{\hspace{0.5cm}} ccc@{\hspace{0.5cm}} |@{\hspace{0.5cm}} cc}
		\begin{tikzpicture}[scale=0.6]
			\node[vertexs] (S) {$S$};
			\node[vertext] (T)  [below=of S] {$T$};
			\node[vertexb] (C)  [right=of S] {$C$};
			\node[vertexb] (B)  [below=of C] {$B$};
			\draw[arrow] (S) -- (T) ;
		\end{tikzpicture} & \quad & \quad &
		\begin{tikzpicture}[scale=0.6]
			\node[vertexs] (S) {$S$};
			\node[vertext] (T)  [below=of S] {$T$};
			\node[vertexb] (C)  [right=of S] {$C$};
			\node[vertexb] (B)  [below=of C] {$B$};
			\draw[arrow] (S) -- (B) node [midway, above] {};
		\end{tikzpicture} & \quad & \quad &

		\begin{tikzpicture}[scale=0.6]
			\node[vertexs] (S) {$S$};
			\node[vertext] (T)  [below=of S] {$T$};
			\node[vertexb] (C)  [right=of S] {$C$};
			\node[vertexb] (B)  [below=of C] {$B$};
			\draw[arrow] (S) -- (C) node [midway, above] {};
		\end{tikzpicture} 
	\end{tabular}
\end{center}
\caption{Atomic SPL graphs: $A_\epsilon$ (left), $A_{\texttt{break}}$ (middle), and $A_{\texttt{continue}}$ (right)~\cite{RegisterAllocation}.}
\label{fig:atom}
\end{figure}

\paragraph{Atomic SPL graphs} There are three different atomic SPL graphs: $A_\epsilon$, $A_{\texttt{break}}$, and $A_{\texttt{continue}},$ corresponding to the programs $\epsilon, \texttt{break}$ and $\texttt{continue},$ respectively. All of them contain only the four special nodes and only one edge as shown in Figure~\ref{fig:atom}.
\paragraph{SPL Operations}	SPL defines three operations. Let $G_1 = (V_1, E_1, S_1, T_1,$ $B_1, C_1)$ and 
	$G_2 = (V_2, E_2, S_2, T_2, B_2, C_2)$ be two disjoint SPL graphs. Then, the graphs obtained by the following operations are also SPL graphs.
	\begin{compactenum}
	\item \emph{Series Operation.} $\oseries{G_1}{G_2}$ is generated by taking the union 
	of $G_1$ and $G_2$ and merging the pairs of vertices $M = (T_1, S_2)$,  $B = (B_1, B_2)$, 
	and $C = (C_1, C_2)$. The distinguished vertices of $\oseries{G_1}{G_2}$ are $(S_1, T_2, B, C)$. 
	It is easy to verify that the series operation is associative. Figure~\ref{fig:series} shows two examples 
	of the series operation. Intuitively, if $G_1$ is the CFG of a program $P_1$ and $G_2$ is the CFG of $P_2$ then $\oseries{G_1}{G_2}$ is the CFG of the program $P_1 ; P_2.$
	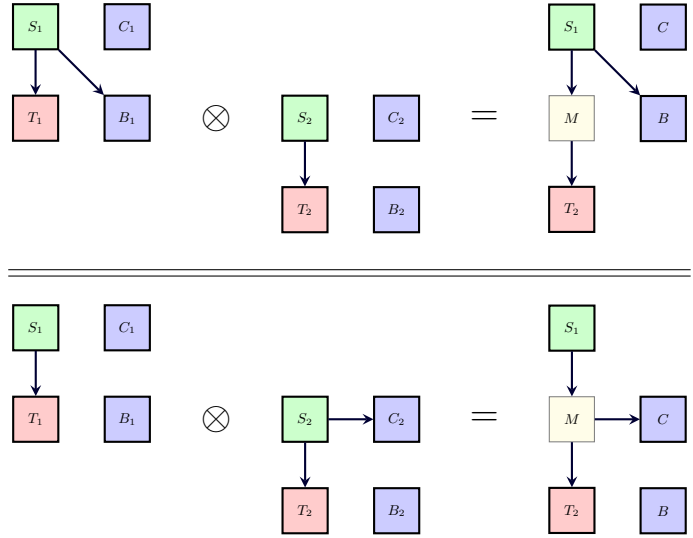
\begin{figure}[H]
		\begin{center}
			\begin{tabular}{cc|ccc|cc}
				\begin{tikzpicture}[scale=0.6]
					\node[vertexs] (S1) {$S_1$};
					\node[vertext] (T1)  [below=of S1] {$T_1$};
					\node[vertexb] (C1)  [right=of S1] {$C_1$};
					\node[vertexb] (B1)  [below=of C1] {$B_1$};
					\draw[arrow] (S1) -- (T1) node [midway, left] {};
					\draw[arrow] (S1) -- (B1) node [midway, above] {};
					\node[emptynode] (R1) [below=of B1] {};
					
					\node (z) [right=of B1] {\fontsize{52}{58}\sffamily\bfseries$\oseries{}{}$};
					
					\node[vertexs] (S2) [right=of z] {$S_2$};
					\node[vertext] (T2)  [below=of S2] {$T_2$};
					\node[vertexb] (C2)  [right=of S2] {$C_2$};
					\node[vertexb] (B2)  [below=of C2] {$B_2$};
					\draw[arrow] (S2) -- (T2) ;
					
					\node (e) [right=of C2] {\fontsize{52}{58}\sffamily\bfseries$=$};
					
					\node[whitevertex] (M) [right=of e]{$M$};
					\node[vertexs] (S) [above=of M] {$S_1$};
					\node[vertext] (T)  [below=of M] {$T_2$};
					\node[vertexb] (C)  [right=of S] {$C$};
					\node[vertexb] (B)  [below=of C] {$B$};
					\draw[arrow] (S) -- (M) ;
					\draw[arrow] (S) -- (B) ;
					\draw[arrow] (M) -- (T) ;
				\end{tikzpicture} \\ \\ \hline \hline \\
				\begin{tikzpicture}[scale=0.6]
					\node[vertexs] (S1) {$S_1$};
					\node[vertext] (T1)  [below=of S1] {$T_1$};
					\node[vertexb] (C1)  [right=of S1] {$C_1$};
					\node[vertexb] (B1)  [below=of C1] {$B_1$};
					\draw[arrow] (S1) -- (T1) ;
					\node[emptynode] (R1) [below=of B1] {};
					
					\node (z) [right=of B1] {\fontsize{52}{58}\sffamily\bfseries$\oseries{}{}$};
					
					\node[vertexs] (S2) [right=of z] {$S_2$};
					\node[vertext] (T2)  [below=of S2] {$T_2$};
					\node[vertexb] (C2)  [right=of S2] {$C_2$};
					\node[vertexb] (B2)  [below=of C2] {$B_2$};
					\draw[arrow] (S2) -- (T2) node [midway, left] {};
					\draw[arrow] (S2) -- (C2) node [midway, above] {};
					
					\node (e) [right=of C2] {\fontsize{52}{58}\sffamily\bfseries$=$};
					
					\node[whitevertex] (M)  [right=of e] {$M$};
					\node[vertexs] (S) [above=of M]{$S_1$};
					\node[vertext] (T)  [below=of M] {$T_2$};
					\node[vertexb] (C)  [right=of M] {$C$};
					\node[vertexb] (B)  [below=of C] {$B$};
					\draw[arrow] (S) -- (M);
					\draw[arrow] (M) -- (T);
					\draw[arrow] (M) -- (C);
				\end{tikzpicture}
				
			\end{tabular}
		\end{center}
		\caption{Two examples of the series operation $\oseries{}{}$, taken from~\cite{RegisterAllocation}.}
		\label{fig:series}
	\end{figure}
	\item \emph{Parallel Operation.} $\oparal{G_1}{G_2}$ is generated by taking union of $G_1$ and $G_2$ and 
	merging the pairs of vertices $S = (S_1, S_2)$, $T = (T_1, T_2)$, $B = (B_1, B_2)$, and $C = (C_1, C_2)$. The 
	special vertex tuple of $\oseries{G_1}{G_2}$ is $(S, T, B, C).$ Figure~\ref{fig:parallel} shows an example 
	of this operation. Informally, if $G_1$ and $G_2$ are the CFGs of the programs $P_1$ and $P_2,$ respectively, then $\oparal{G_1}{G_2}$ is the CFG of the program  $\texttt{if} ~\varphi~ \texttt{then} ~P_1~ \texttt{else} ~P_2~ \texttt{fi}.$
			\begin{figure}[H]
		\begin{center}
			\begin{tikzpicture}[scale=0.6]
				\node[whitevertex] (M1) {$M_1$};
				\node[vertexs] (S1) [above=of M1] {$S_1$};
				\node[vertext] (T1)  [below=of M1] {$T_1$};
				\node[vertexb] (B1)  [right=of S1] {$B_1$};
				\node[vertexb] (C1)  [below=of B1] {$C_1$};
				\draw[arrow] (S1) -- (M1) ;
				\draw[arrow] (S1) -- (B1) ;
				\draw[arrow] (M1) -- (T1) ;
				
				\node (z) [right=of C1] {\fontsize{52}{58}\sffamily\bfseries$\oparal{}{}$};
				
				\node[whitevertex] (M2) [right=of z]  {$M_2$};
				\node[vertexs] (S2) [above=of M2]{$S_2$};
				\node[vertext] (T2)  [below=of M2] {$T_2$};
				\node[vertexb] (C2)  [right=of M2] {$C_2$};
				\node[vertexb] (B2)  [above=of C2] {$B_2$};
				\draw[arrow] (S2) -- (M2);
				\draw[arrow] (M2) -- (T2);
				\draw[arrow] (M2) -- (C2);
				
				\node (e) [right=of C2] {\fontsize{52}{58}\sffamily\bfseries$=$};
				
				\node[whitevertex] (aM1) [right=of e] {$M_1$};
				\node[vertexs] (aS) [above right=of aM1] {$S$};
				\node[whitevertex] (aM2) [below right=of aS] {$M_2$};
				\node[vertext] (aT)  [below left=of aM2] {$T$};
				\node[vertexb] (aC)  [right=of aM2] {$C$};
				\node[vertexb] (aB)  [right=of aS] {$B$};
				\draw[arrow] (aS) -- (aM1);
				\draw[arrow] (aS) -- (aM2);
				\draw[arrow] (aS) -- (aB);
				\draw[arrow] (aM1) -- (aT);
				\draw[arrow] (aM2) -- (aT);
				\draw[arrow] (aM2) -- (aC);
			\end{tikzpicture}
		\end{center}
		\caption{An example of the parallel operation $\oparal{}{}$~\cite{RegisterAllocation}.}
		\label{fig:parallel}
		\end{figure}
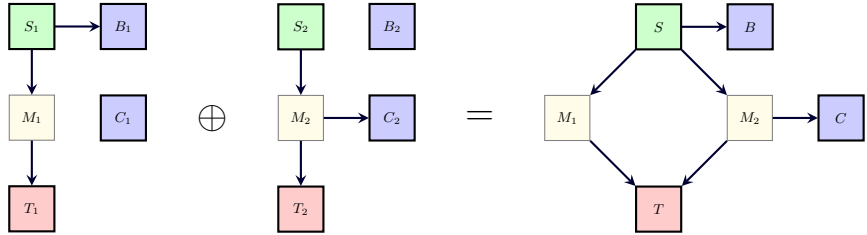
	\item \emph{Loop Operation.} $\oloop{G_1}$ is generated by adding four new vertices $S, T, B, C$ to $G_1$ and 
	then adding the following edges: $(S, S_1), (S, T), (T_1, S), (C_1, S),$ and $(B_1, T).$ The special vertex 
	tuple of $\oloop{G_1}$ is $(S, T, B, C).$ Figure~\ref{fig:loop} shows an example of the loop operation. Intuitively, if $G_1$ is the CFG of the program $P_1,$ then $\oloop{G_1}$ is the CFG of the loop with $P_1$ as its body, i.e.~$\texttt{while} ~\varphi~ \texttt{do} ~P_1~ \texttt{od}.$
		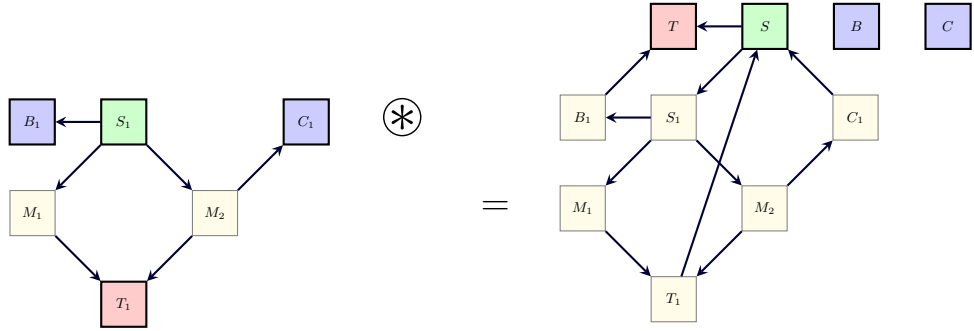
\begin{figure}[H]
		\begin{center}
			\begin{tikzpicture}[scale=0.6]
				\node[whitevertex] (aM1)  {$M_1$};
				\node[vertexs] (aS) [above right=of aM1] {$S_1$};
				\node[whitevertex] (aM2) [below right=of aS] {$M_2$};
				\node[vertext] (aT)  [below left=of aM2] {$T_1$};
				\node[vertexb] (aC)  [above right=of aM2] {$C_1$};
				\node[vertexb] (aB)  [left=of aS] {$B_1$};
				\draw[arrow] (aS) -- (aM1);
				\draw[arrow] (aS) -- (aM2);
				\draw[arrow] (aS) -- (aB);
				\draw[arrow] (aM1) -- (aT);
				\draw[arrow] (aM2) -- (aT);
				\draw[arrow] (aM2) -- (aC);
				
				\node (star) [right=of aC] {\fontsize{52}{58}\sffamily\bfseries$\oloop{}$};
				
				\node (e) [below right=of star] {\fontsize{52}{58}\sffamily\bfseries$=$};
				
				\node[whitevertex] (baM1) [right=of e] {$M_1$};
				\node[whitevertex] (baS) [above right=of baM1] {$S_1$};
				\node[vertexs] (S) [above right=of baS] {$S$};
				\node[vertexb] (B) [right=of S] {$B$};
				\node[vertexb] (C) [right=of B] {$C$};
				\node[whitevertex] (baM2) [below right=of baS] {$M_2$};
				\node[whitevertex] (baT)  [below left=of baM2] {$T_1$};
				\node[whitevertex] (baC)  [above right=of baM2] {$C_1$};
				\node[whitevertex] (baB)  [left=of baS] {$B_1$};
				\node[vertext] (T) [left=of S] {$T$};
				\draw[arrow] (baS) -- (baM1);
				\draw[arrow] (baS) -- (baM2);
				\draw[arrow] (baS) -- (baB);
				\draw[arrow] (baM1) -- (baT);
				\draw[arrow] (baM2) -- (baT);
				\draw[arrow] (baM2) -- (baC);
				\draw[arrow] (S) -- (baS);
				\draw[arrow] (S) -- (T);
				\draw[arrow] (baT) -- (S);
				\draw[arrow] (baB) -- (T);
				\draw[arrow] (baC) -- (S);
			\end{tikzpicture}
		\end{center}
		\caption{An example of the loop operation $\oloop{}$~\cite{RegisterAllocation}.}
		\label{fig:loop}
	\end{figure}
\end{compactenum}
We say an SPL graph $G=(V, E, S, T, B, C)$ is \emph{closed} if there are no incoming edges to the vertices $B$ and $C.$

\paragraph{SPLs as CFGs} Given the above definitions of structured programs and SPL graphs, we have the following homomorphism which maps every structured program to its control-flow graph. Moreover, this homomorphism preserves closedness, i.e.~closed programs are mapped to closed graphs. A graph is an SPL graph if and only if it is the control-flow graph of a program~\cite{RegisterAllocation}. 
$$
\begin{matrix}
\cfg{\epsilon} = A_\epsilon & \cfg{\texttt{break}} = A_{\texttt{break}} & \cfg{\texttt{continue}} = A_{\texttt{continue}}
\end{matrix}
$$
$$
	\cfg{P_1 ; P_2} = \oseries{\cfg{P_1}}{\cfg{P_2}}
$$
$$
	\cfg{\texttt{if} ~\varphi~ \texttt{then} ~P_1~ \texttt{else} ~P_2~ \texttt{fi}} = \oparal{\cfg{P_1}}{\cfg{P_2}}
$$
$$
\cfg{\texttt{while} ~\varphi~ \texttt{do} ~P_1~ \texttt{od}} = \oloop{\cfg{P_1}}
$$
\paragraph{SPL Decompositions} Given a closed program $P,$ we can first parse it based on the grammar~\eqref{gram:prog} to generate a parse tree. Subsequently, by 
applying the homomorphism above to this parse tree, we can derive a parse tree according to~\eqref{gram:spl} for its control-flow graph. We use the term \emph{SPL decomposition} to refer to the parse tree of the CFG according to~\eqref{gram:spl}. It is easy to verify that this process takes linear time. See Figure~\ref{fig:decompo} as an example.
\begin{figure}
	\begin{subfigure}{0.3\textwidth}
		\begin{lstlisting}[mathescape,numbers=none]
while $x \geq 1$ do
	if $x \geq y$ $\text{then}$
		$x \gets x - y;$
		$\text{break}$
	else
		$y \gets y - x;$
		$\text{continue}$
	fi
od
		\end{lstlisting} 
	\end{subfigure} \hspace{3cm}
	\begin{subfigure}{0.5\textwidth}
		\begin{tikzpicture}[scale=0.6]
			\node[rect] (eps1) {$\epsilon$};
			\node[rect] (break) [right=of eps1] {$\texttt{break}$};
			\node[rect] (eps2) [right=of break] {$\epsilon$};
			\node[rect] (continue) [right=of eps2] {$\texttt{continue}$};
			\node[rect] (sem1) [above=of eps1] {$;$};	
			\node[rect] (sem2) [above=of eps2] {$;$};	
			\node[rect] (if) [above=of sem1] {$\texttt{if}$};	
			\node[rect] (while) [above=of if] {$\texttt{while}$};	
			\draw[thick] (while) -- (if) ;
			\draw[thick] (if) -- (sem1) ;
			\draw[thick] (if) -- (sem2) ;
			\draw[thick] (sem1) -- (eps1) ;
			\draw[thick] (sem1) -- (break) ;
			\draw[thick] (sem2) -- (eps2) ;
			\draw[thick] (sem2) -- (continue) ;
		\end{tikzpicture}
	\end{subfigure}
	
	\begin{subfigure}{0.33\textwidth}
		\begin{tikzpicture}[scale=0.6]
			\node[rect] (eps1) {$A_\epsilon$};
			\node[rect] (break) [right=of eps1] {$A_\texttt{break}$};
			\node[rect] (eps2) [right=of break] {$A_\epsilon$};
			\node[rect] (continue) [right=of eps2] {$A_\texttt{continue}$};
			\node[rect] (sem1) [above=of eps1] {$\oseries{}{}$};	
			\node[rect] (sem2) [above=of eps2] {$\oseries{}{}$};	
			\node[rect] (if) [above=of sem1] {$\oparal{}{}$};	
			\node[rect] (while) [above=of if] {$\sloop$};	
			\draw[thick] (while) -- (if) ;
			\draw[thick] (if) -- (sem1) ;
			\draw[thick] (if) -- (sem2) ;
			\draw[thick] (sem1) -- (eps1) ;
			\draw[thick] (sem1) -- (break) ;
			\draw[thick] (sem2) -- (eps2) ;
			\draw[thick] (sem2) -- (continue) ;
		\end{tikzpicture}
	\end{subfigure} \hspace{2.3cm}
	\begin{subfigure}{0.33\textwidth}
		\begin{tikzpicture}[scale=0.6,node distance=1cm and 2cm]
			
			\node[whitevertex] (baM1)  {$M_1$};
			\node[whitevertex] (baB)  [left=of baM1] {$B_1$};
			\node[whitevertex] (baS) [right=of baM1] {$S_1$};
			\node[whitevertex] (baM2) [right=of baS] {$M_2$};
			\node[vertext] (T) [above=of baM1] {$T$};
			
			\node[vertexs] (S) [above=of baS] {$S$};
			\node[whitevertex] (baC)  [above=of baM2] {$C_1$};

			\node[vertexb] (C) [below=of baS] {$C$};
			\node[vertexb] (B) [left=1cm of C] {$B$};
			
			\node[whitevertex] (baT)  [right=1cm of C] {$T_1$};

			\draw[arrow] (S) -- (T) node [midway, above] {$x<1$};
			\draw[arrow] (S) -- (baS) node [midway, left] {$x\geq 1$};
			\draw[arrow] (baS) -- (baM1) node [midway, above] {$x \geq y$};
			\draw[arrow] (baS) -- (baM2) node [midway, above] {$x < y$};
			\draw[arrow] (baM1) -- (baB) node [midway, above] {$x \gets x - y$};
			\draw[arrow] (baM2) -- (baC) node [midway, right] {$y \gets y - x$};
			\draw[arrow] (baB) -- (T) node [midway, left] {$\texttt{break}$};
			\draw[arrow] (baC) -- (S) node [midway, above] {$\texttt{continue}$};
			\draw[arrow] (baT) -- (S);
		\end{tikzpicture}
	\end{subfigure}
	\caption{A program $P$ (top left), its parse tree (top right), the corresponding parse tree of $G = \cfg{P}$ (bottom left) and the graph $G = \cfg{P}$ (bottom right)~\cite{RegisterAllocation}. The edges of the graph are labeled according to the statements in the program.}
	\label{fig:decompo}
\end{figure}

%% file: sections/PCSP.tex
\section{Our Algorithm} \label{sec:pcsp}

\begin{figure}
    \centering
    \subfloat[gragh]{
   \begin{tikzpicture}[scale=0.6]
    \node[whitevertex] (A) at (0, 1.5) {A};
    \node[whitevertex] (B) at (-1.5, 0) {B};
    \node[whitevertex] (C) at (1.5, 0) {C};
    \node[whitevertex] (D) at (0, -1.5) {D};

    \draw (A) -- (B);
    \draw (A) -- (C);
    \draw (B) -- (D);
    \draw (C) -- (B);
\end{tikzpicture}
}
    \subfloat[three colors]{
   \begin{tikzpicture}[scale=0.6]
    \node[vertexs] (A) at (0, 1.5) {A};
    \node[vertext] (B) at (-1.5, 0) {B};
    \node[vertexb] (C) at (1.5, 0) {C};
    \node[vertexs] (D) at (0, -1.5) {D};

    \draw (A) -- (B);
    \draw (A) -- (C);
    \draw (B) -- (D);
    \draw (C) -- (B);
\end{tikzpicture}
}
 \subfloat[two colors]{
   \begin{tikzpicture}[scale=0.6]
    \node[vertexs] (A) at (0, 1.5) {A};
    \node[vertexb] (B) at (-1.5, 0) {B};
    \node[vertexb] (C) at (1.5, 0) {C};
    \node[vertexs] (D) at (0, -1.5) {D};

    \draw (A) -- (B);
    \draw (A) -- (C);
    \draw (B) -- (D);
    \draw (C) -- (B);
\end{tikzpicture}
}
    \caption{Graph Coloring}
    \label{fig:gc}
\end{figure}
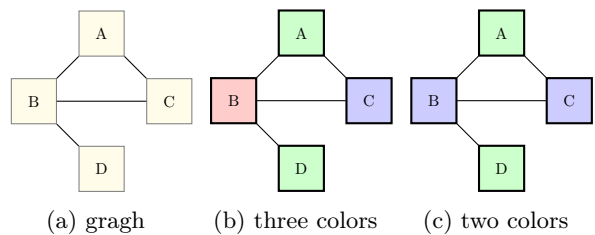

\paragraph{CSP} The well-known Constraint Satisfaction Problem (CSP) framework~\cite{PCS} defines a tuple \(\langle V, D, C \rangle\), where \(V\) is a finite set of variables, \(D\) is the domain set for all \(v \in V\), and \(C\) is a set of constraints\footnote{Our algorithm can be trivially extended to support different domains for variables.}. A CSP is solved by finding an assignment of values to the variables such that all constraints are satisfied.  When applied to graphs, we treat each node as a variable and each edge as a constraint, with the stipulation that constraints exist only between adjacent nodes. For example, the graph coloring problem can be formulated as a CSP. Considering the graph~\ref{fig:gc}, given three colors, we want to color the graph so that no two adjacent nodes share the same color. We can consider each node to be a variable, and each edge imposes a constraint that the colors of adjacent nodes must differ, then this graph coloring problem is a typical CSP. It is well-known that the graph coloring problem is NP-hard even when limiting the domain set to only three colors, which implies that the CSP problem is also NP-hard.

\paragraph{PCSP} In the context of graph-based PCSPs (Partial Constraint Satisfaction Problems)~\cite{PCS}, we allow certain constraints to be violated at a specified cost, with the goal of finding a solution that minimizes this cost. To define the cost, we introduce a cost function \(c(e, b_0, b_1)\), where \(e\) is the edge, and \(b_0\) and \(b_1\) are the values assigned to the two nodes connected by the edge. If \(b_0\) and \(b_1\) do not violate the constraints, the cost is 0; otherwise, a positive cost is assigned. Our objective is to find:
\[
\min_{A} \sum_{e \in E} c(e, A(v_0), A(v_1))
\]
where \(A: V \to D\) is a valuation that maps each node to a domain element.

Figure~\ref{fig:gc} with two colors is an example. Similar to CSP, we consider each node as a variable and each color as a value. We define the cost function as follows:

\[
c(e, b_0, b_1) = 
\begin{cases} 
0 & \text{if } b_0 \neq b_1 \\ 
1 & \text{if } b_0 = b_1 
\end{cases} 
\]

In this scenario, the minimum cost is 1, as the edge connecting nodes B and C incurs a cost of 1, while all other edges do not contribute to the cost. It is straightforward to verify that this configuration constitutes a valid solution to the PCSP problem, as it is impossible to color the figure using only two colors in a manner that satisfies the constraints.
If we assign an infinite cost to the constraints, the problem reduces to the CSP. Thus, PCSP is NP-hard, too.

\paragraph{Our Setting} We consider a PCSP instance in which the underlying graph is the control-flow graph of a structured goto-free program. Our goal is to use the SPL decomposition of the control-flow graph to obtain an efficient algorithm. As we will see further below, this is motivated by the fact that many classical problems in program analysis and compiler optimization are defined over the control-flow graphs and are special cases of PCSP.

\paragraph{Our SPL-based Algorithm} Our algorithm proceeds with a bottom-up dynamic programming on the SPL decomposition. Note that each node $u$ of the SPL decomposition corresponds to an SPL subgraph $G_u = (V_u, E_u, S_u, T_u, B_u, C_u)$ of $G$ which is either an 
atomic SPL graph (when $u$ is a leaf) or obtained by applying one of the SPL operations to the graphs corresponding to the children of $u.$
Let $\varGamma_u = \{S_u, T_u, B_u, C_u\}$ be the set of special vertices of $G_u$. Let $X: \varGamma_u \rightarrow D$ be the assignment for the vertices in $\varGamma_u$. We 
define a dynamic programming variable $\DP[u, X].$ Our goal is to compute this dynamic programming value such that
$$
\DP[u, X] = \min_{A}\sum_{e \in E_u} c(e, A(v_0), A(v_1))
$$
where the minimum is taken over all assignments $A$ to the vertices of $G_u$ that agree with $X$ on the special vertices. In other words, $A(v) = X(v)$ for all $v \in \varGamma_u$. We now show how our algorithm computes these $\DP[\cdot, \cdot]$ values at every kind of node in the SPL decomposition:

\begin{enumerate}
    \item \emph{Atomic Nodes}: If $G_u$ is an atomic SPL graph, then the only vertices in $G_u$ are the four special vertices. Therefore, we must have $A=X.$ Our algorithm sets
	 $$\DP[u, X] = \sum_{e \in E_u} c(e, X(v_0), X(v_1))$$

    \item \emph{Series Nodes}: Suppose $G_u = \oseries{G_v}{G_w}$ where $v$ and $w$ are the children of $u$ in the SPL decomposition. Let $X_i$ be the assignment for the special vertices/variables in $\varGamma_i$ for $i \in \{u, v, w\}.$  
     We say that $X_u$ and $X_w$ are compatible and write $X_u \compatible X_w$ if the following conditions satisfied:
    \begin{itemize}
    \item 	$X_u(S_u)=X_v(S_v);$
    \item	$X_u(B_u)=X_v(B_v);$
    \item	$X_u(C_u)=X_v(C_v).$
    \end{itemize}
    Intuitively, compatibility means that the assignments $X_u$ and $X_v$ return the same value when given the same vertex/variable.

    Now consider $X_u$ and $X_w$. We say that $X_u$ and $X_w$ are compatible and write $X_u \compatible X_w$ if the following conditions are satisfied:
    \begin{itemize}
        \item 	$X_u(T_u)=X_w(T_w);$
        \item	$X_u(B_u)=X_w(B_w);$
        \item	$X_u(C_u)=X_w(C_w).$
        \end{itemize}
    The intuition is the same as the previous case, except that we now have $T_u = T_w.$ Finally, we say that $X_v$ and $X_w$ are compatible and write $X_v \compatible X_w$ if 
    \begin{itemize}
        \item $X_v(T_v)=X_w(S_w).$
    \end{itemize}
    This is because $T_v$ and $S_w$ are the same vertex of the CFG.
    
    Given the definition above, our algorithm simply sets
    $$
	\DP[u, X_u] = \min_{\makecell[c]{X_u \compatible X_v \\ X_u \compatible X_w \\ X_v \compatible X_w}} \DP[v, X_v] + \DP[w, X_w].
    $$
    This is because every edge in $G_u$ appears in either $G_v$ or $G_w$ but not both. Thus, the cost of the edges would simply be the sum of their costs in the two subgraphs.

    \item \emph{Parallel Nodes:} We can handle parallel nodes in the same manner as series nodes, i.e., ~finding compatible assignments at both children. 
    To be more precise, let $G_u = \oparal{G_v}{G_w}.$ The compatibility conditions are now changed to simple equality. This is because the parallel operation merges the respective start, terminate, break and continue vertices in $G_v$ and $G_w.$ In some cases, repeated edges may be introduced, e.g., both graphs \( v \) and \( w \) have edges connecting the \( S \) vertex and the \( B \) vertex. These two edges represent the same edge in graph \( u \), so we only need to calculate the cost once. Let's create a set \( E' \) that contains all such edges.

    With the same argument as in the previous case, our algorithm sets
    $$
    \DP[u, X_u] = \min_{\makecell[c]{X_u \compatible X_v \\ X_u \compatible X_w \\ X_v \compatible X_w}} \DP[v, X_u] + \DP[w, X_u] - \sum_{e\in E'}{c(e)}.
    $$

    \item \emph{Loop Nodes:} Finally, we should handle the case where $G_u = \oloop{G_v}.$ 
    By construction, in comparison to $G_v,$ the graph $G_u$ has four new vertices 
    $$\varGamma_u=\{S_u, T_u, B_u, C_u\}$$ and three new edges 
    $$E_{\text{new}} = \{ (S_u, S_v), (S_u, T_u), (T_v, S_u)) \}.$$ 
    Knowing $X_u$ and $X_v$ is sufficient to calculate the costs of our new edges. Let us denote the cost of a new edge $e$ based on $X_u$ and $X_v$ by $c(e)$.
  Our algorithm sets
    $$
    \DP[u, X_u] = \min \DP[v, X_v] + \sum_{e \in E_{\text{new}}}c(e).
    $$
\end{enumerate}

After computing all the $\DP[\cdot, \cdot]$ values, our solution, i.e.,~the minimum possible cost, is given by $\min_X \DP[r, X]$ 
where $r$ is the root node of the SPL decomposition and $X : \varGamma_r \rightarrow D$ is an assignment for the special vertices of the root SPL graph. The exact
assignment for every node can be easily tracked and updated during the dynamic programming.


 Let us now analyze the time complexity of each step of our algorithm:
\begin{itemize}
    \item \emph{Atomic nodes:} As we only have four vertices in an atomic graph and each of them have $|D|$ possible values,
    the time complexity is \(O(|D|^4)\). To slightly optimize the runtime, we can only consider the connected vertices and ignore disconnected vertices, which, by definition, have no constraints on them. As there are at most two connected
    vertices in atomic nodes, the time complexity is \(O(|D|^2)\). 
    \item \emph{Series nodes:} When doing computations for a series node, there are a total of eight special vertices that need to be considered, but as we also need to 
    consider the compatibility, there are three pairs among them that are guaranteed to have the same values. Thus, the time complexity is
    \(O(|D|^5)\), as we only need to consider value for five independent variables. 
    \item \emph{Parallel nodes:} The analysis is similar to a series node, except that there are four pairs of vertices that need to have the same value. Thus,
     the time complexity is $O(|D|^4).$
    \item \emph{Loop nodes:} This is also similar to the previous cases, but no vertices are merged and should have the same value.
    Thus, the time complexity is $O(|D|^8)$. However, we know that the $C_u$ and $B_u$ are not connected to any other vertex after the loop operation.
   So, we can temporarily ignore them, reducing the time complexity to $O(|D|^6)$.
\end{itemize}
Thus, the dynamic programming computations at each node of the SPL decomposition can be done in $O(|D|^6).$ Moreover, the SPL decomposition has the same asymptotic size as the control-flow graph. Therefore, the overall runtime of our algorithm is in $O(|G|\cdot|D|^6)$. 

Before discussing the placement of the bank selection problem, let's first present how two other applications~\cite{cai2024faster,RegisterAllocation} that have utilized SPL-decomposition can be considered as instances of the PCSP.

\paragraph{Register Allocation}
Here we show how Register Allocation~\cite{RegisterAllocation} is a special case of PCSP. Suppose we are given a program $P$ with control-flow graph $G=\cfg{P} = (V, E, S, T, B, C).$ Let $[r] = \{0, 1, \ldots, r-1\}$ be the set of available registers and $\vars$ the set of our program variables. Every variable $v \in \vars$ has a lifetime $\lt{v}$ which is a connected subgraph of $G.$ Since lifetimes can be computed by a simple data-flow analysis, we assume without loss of generality that they are given as inputs to our algorithm. For a vertex $v$ or edge $e$ of $G,$ we denote the set of variables that are alive at this vertex/edge by $L(v)$ or $L(e).$ An \emph{assignment} is a function $f: \vars \rightarrow [r] \cup \{\perp\}$ which maps each variable either to a register or to $\perp.$ The latter models the variable being spilled. An assignment is valid if it does not map two variables with intersecting lifetimes to the same register. We denote the set of all valid assignments by $F.$ . Cost is introduced when we need to switch the value of a register, and the goal is to find the assignment with minimum cost.

In this case, the "value" assigned to each node is the allocation of an alive variable. Suppose there are at most $V$ alive variables at one node and there are $r$ registers, then we can create a completed inference graph with $V$ nodes and try to color them with $r$ colors. Then the domain size is $$\binom{V}{r} \cdot r! + \binom{V}{r-1} \cdot (r-1)! + \dots + \binom{V}{0} \cdot 0! \in O(r \cdot V^r).$$ Hence with our algorithm, the time complexity is $O(|G| \cdot r^6 \cdot V^{6 \cdot r}).$ 

\paragraph{LOSPRE}
LOSPRE~\cite{cai2024faster} is another example of the PCSP. It is a modern redundancy elimination technique that calculates repeated expressions only once and stores the results in a template variable. Each time we need to use that expression, we can directly utilize the template variable instead of recalculating it. With the control flow graph \( G = \{V, E\} \), we can define this problem as follows:

\begin{itemize}
    \item \emph{Use set:} Consider an expression $e.$ We define the use set $U$ of $e$ as the set of all nodes of the CFG in which the expression $e$ is computed.

    \item \emph{Life set:} Our goal is to precompute the expression $e$ at a few points, save the result in a temporary variable \texttt{temp}, and then use   \texttt{temp} in place of $e$ in every node of $U.$ We denote the lifetime of the variable $\texttt{temp}$ by $L$ and call it our life set. 

    \item \emph{Invalidating set:} We say a node $v$ of the CFG invalidates $e$ if the statement at $v$ changes the value of $e.$ For example, if $e = \texttt{a+b},$ then the statement $\texttt{a = 0}$ invalidates $e.$ We denote the set of all invalidating nodes by $I.$ These nodes play a crucial role in LOSPRE since they force us to update the value saved in $\texttt{temp}$ by recomputing $e.$ We assume that the entry and exit nodes are invalidating since LOSPRE is an intraprocedural analysis that has no information about the program's execution before or after the current function.

    \item \emph{Calculating set:} Given the sets $U, L$ and $I$ above, we have to make sure the value of our temporary variable $\texttt{temp}$ is correct at every node in $U \cup L.$ Thus, for every edge $(x, y) \in E$ of the CFG where $x \not \in L$ and $y \in U \cup L,$ we have to insert a computation $\texttt{temp} = e$ between $x$ and $y.$ Similarly, if $x \in I,$ then the value stored at $\texttt{temp}$ becomes invalid after the execution of $x,$ requiring us to inject the same computation between $x$ and $y.$ Formally, the computation $\texttt{temp} = e$ has to be injected into the following set of edges of the CFG:
$$
C(U, L, I) = \{ (x, y) \in E ~\vert~ x \not\in L \setminus I ~\land~ y \in U \cup L \}.
$$
\end{itemize}

This time, there are two types of costs associated with the process above: (i)~injecting calculations into the edges in $C(U, L, I)$ and (ii)~keeping an extra variable $\texttt{temp}$ at every node in $L.$ These costs are dependent on the goals pursued by the compiler. For example, a compiler aiming to minimize code size will focus on (i). On the other hand, if our goal is to ease register pressure, we would want to minimize (ii). LOSPRE is an expressive framework in which these costs are modeled by two functions
$$
c: E \rightarrow K
$$ 
and 
$$
l: V \rightarrow K.
$$
where $K$ is a totally-ordered set with an addition operator, $c$ is a function that maps each edge to the cost of adding a computation of $e$ in that edge and $l$ is similarly a function that maps each vertex of the CFG to the cost of keeping the temporary variable $\texttt{temp}$ alive at that vertex. Our goal is to find a life set $L$ that minimizes the total cost $$\cost(G, U, I, L, c, l) = \sum_{e\in C(U,L,I)}c(e)+\sum_{v\in L} l(v).$$

To consider this problem as PCSP, we only need to have a little modification, as the edge cost is the same as the definition in PCSP, and the node cost can be easily added to the total cost. In this case, the "value" assigned to each node if it is belong to Use set, Life set, and Invalidating set, as for each set, there are 2 different case, hence the domain size is $2^3=8$, and hence, applying to our PCSP algorithm, this can be down in $O(|G|\cdot 8^6)=O(|G|)$, in this case, thanks to the constant domain size, we can develop a linear algorithm without taken any parameter.

%% file: sections/bankSelection.tex
\section{Optimal Placement of Bank Selection Instructions} \label{sec:bank}

As mentioned in the previous section, register allocation and LOSPRE, which were both previously solved using tree decompositions and SPL decompositions are special cases of PCSP. In this section, we cover another classical use-case of PCSP over control-flow graphs, i.e.~that of optimal placement of bank selection instructions. To the best of our knowledge, there are no previous solutions to this problem that exploit SPL decompositions.

\paragraph{Motivation} Partitioned memory architectures are prevalent in 8-bit and 16-bit microcontrollers. In these systems, a portion of the 
logical address space serves as a window into a larger physical address space. The segments of the physical address space 
that can be mapped into this window are referred to as memory banks. A mechanism exists to determine which part of the 
physical address space is visible within the window, typically achieved through bank selection instructions.

The assignment of variables to specific memory banks is generally performed by the programmer, for instance, 
through named address spaces in Embedded C, or by the compiler, using techniques such as bin-packing heuristics to 
minimize RAM usage. Some approaches integrate the placement of variables in memory banks with the insertion of bank 
selection instructions. However, in embedded systems, there is often more available space for code than for data. As a 
result, variables stored in banked memory tend to be larger, making it advantageous to prioritize the efficient packing of 
variables into the banks before addressing other factors such as code size and execution speed. Consequently, 
the placement of variables in memory typically occurs at an earlier stage of the compilation process than the insertion of 
bank-switching instructions.

\paragraph{Domain} The following formal definition of the bank selection problem is adapted from~\cite{bankselection}.  Let $D$ be the memory bank domain, including a special symbol \(\perp \in D\) that indicates that the 
currently selected bank is unknown. A program can be modeled as a control-flow graph \( G = (V, E) \), 
where \( E \subseteq V^2 \), and each node \( v \in V \) can be assigned a memory bank from $D$. Some of the 
nodes are pre-assigned since a specific bank must be active at that node.

\begin{figure}
	\centering
	\subfloat[Program]{
		\includegraphics[clip,width=.3\textwidth]{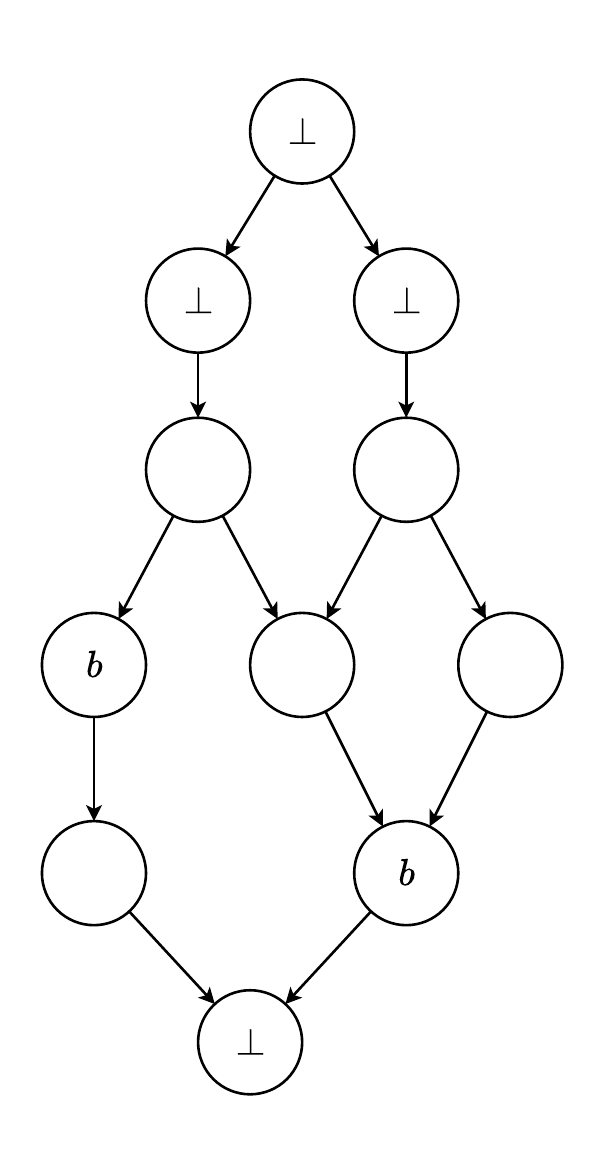}
	}%
	\subfloat[Ad-hoc Approach]{
		\includegraphics[clip,width=.3\textwidth]{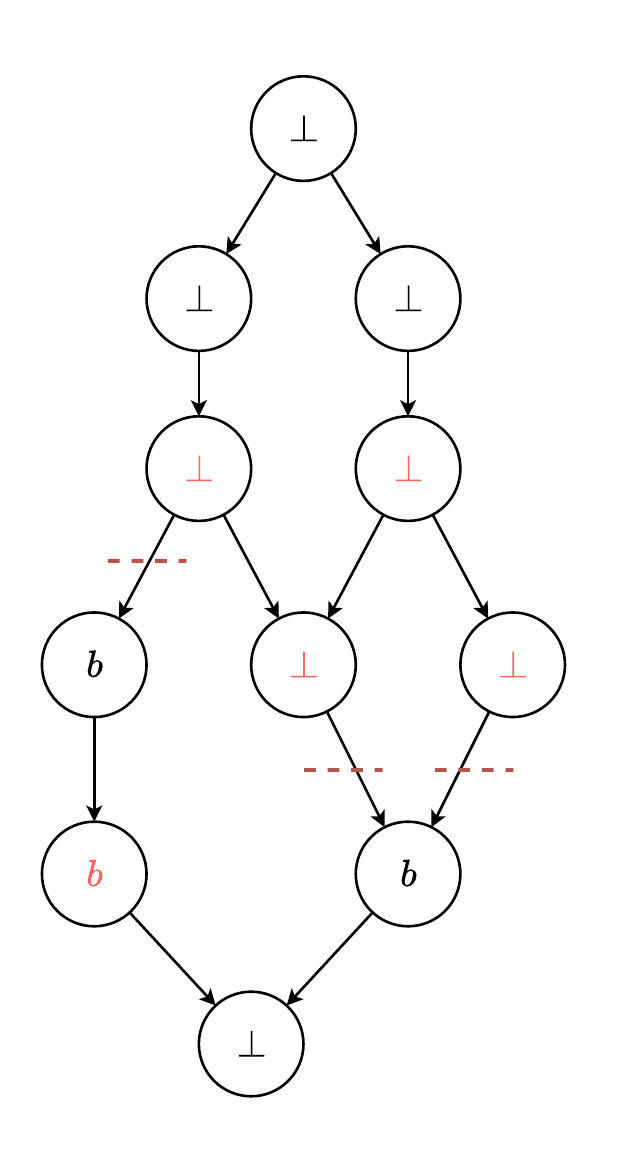}
	}%
	\subfloat[Optimal]{
		\includegraphics[clip,width=.3\textwidth]{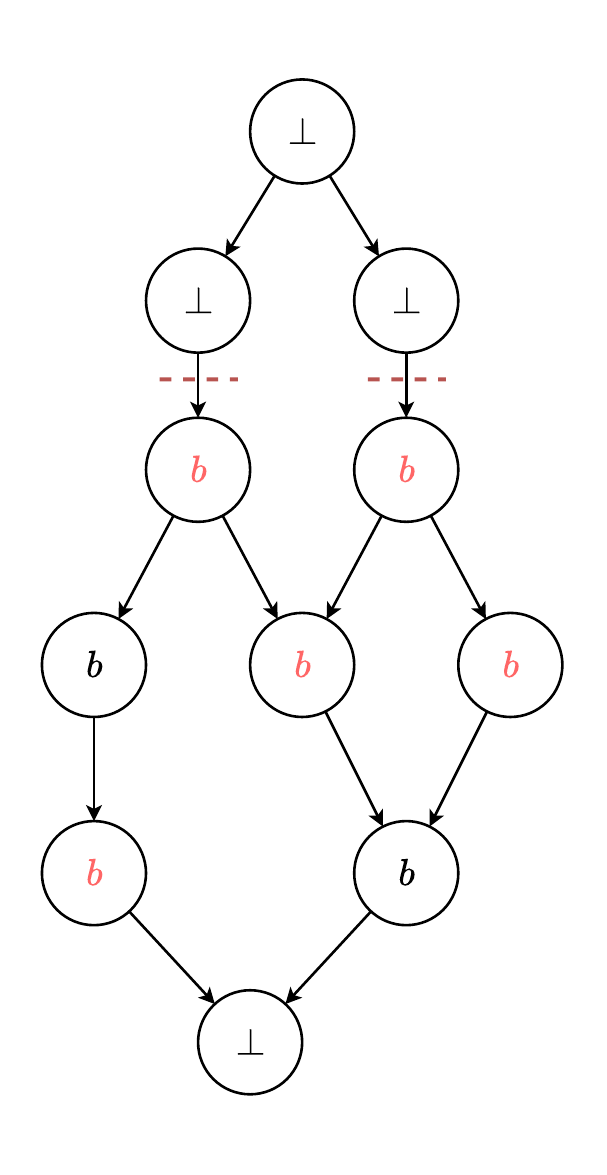}
	}
	\caption{Example Instance}
	\label{fig:bank-selection}
\end{figure}

\paragraph{Example}
Figure~\ref{fig:bank-selection} shows an example of a program with bank selection instructions. The program has two pre-assigned
vertices and the bank $b$ should be active at these vertices. When the program is at other locations, there is no restriction on the selected bank since there is no memory access to any bank.
 As a simple ad-hoc approach, we can add the bank selection instruction just before we need the bank to be active. This is shown in 
 Figure~\ref{fig:bank-selection} (b) and would need the insertion of three bank selection instructions. Suppose we aim to minimize the 
 number of bank selection instructions. Then, we can insert them at the beginning of the program, as shown in Figure~\ref{fig:bank-selection} (c). This only requires two instructions and is the optimal solution. 

\paragraph{Cost Function}
In the classical formulation of the optimal bank selection problem, a cost function is defined over the CFG \( G = (V, E) \). This is a function \( c: E \times D \times D \to \mathbb{R} \) 
that assigns a cost to each edge \( e \in E \) based on the memory banks assigned to its two endpoints. In other words, we incur the cost $c(e,b_0,b_1)$ if $b_0$ is active right before the edge $e$ and $b_1$ becomes active after the edge. Such cost functions may be designed based on different optimization criteria. A typical cost function for optimizing code 
size is as follows: 
\begin{itemize}
    \item \( c(e, b, b) = c(e, b, \perp) = 0 \) since no instructions need to be inserted;
    \item \( c(e, b_0, b_1) = c_1 > 0 \) for \( b_0 \neq b_1 \neq \perp \) when \( e \) is an edge from a taken conditional branch, as splitting such an edge generates an additional unconditional jump instruction. 
    \item For all other cases, \( c(e, b_0, b_1) = c_0 > 0 \) for \( b_0 \neq b_1 \neq \perp \), with the condition that \( c_0 < c_1 \).
\end{itemize}

\paragraph{Objective}
The goal in optimal bank selection is to find an assignment of memory banks to the vertices of the control-flow graph that 
minimizes the total cost of the edges. In other words, we aim to compute
$$
\min_{A} \sum_{e \in E} c(e, A(v_0), A(v_1))
$$
where \( A: V \to D \) maps each node to a domain. We note that this problem is NP-hard, even when the cost function is simplified to \( c(e, b_0, b_1) = 0 \) for \( b_0 = b_1 \) or \( b_1 = \perp \), and \( c(e, b_0, b_1) = 1 \) in all other cases~\cite{bankselection}.

\paragraph{Our Algorithm}
It is easy to verify that optimal insertion of bank selection instructions is an instance of the PCSP graph problem. Thus, we can simply apply the general solution
developed in Section~\ref{sec:pcsp} and have it solved in $O(|G|\cdot|D|^6)$ time, where $|D|$ is the number of possible banks.

%% file: sections/experiments.tex
\section{Experimental Results} \label{sec:exp}

As mentioned in Section~\ref{sec:pcsp}, both register allocation and LOSPRE have previously been solved using SPL decompositions~\cite{cai2024faster,RegisterAllocation}. Indeed, the algorithms provided by~\cite{cai2024faster,RegisterAllocation} can be seen as instances of our general algorithm, when the domain and cost function are fixed as in Section~\ref{sec:pcsp}. Therefore, we inherit their experimental results exactly. In this section, we provide further experimental results over the optimal bank selection problem of Section~\ref{sec:bank}, which, to the best of our knowledge, has not been considered in the context of SPL decompositions before. In addition to that, we also compare our approach with the classical algorithms for NP-hard problems, including SAT and ILP. 

\paragraph{Implementation}
We implemented our bank selection algorithm in \texttt{C++} and integrated it with the Small Device \texttt{C} Compiler (SDCC)~\cite{sdcc1,sdcc2}. We will publish the implementation as free and open-source software dedicated to the public domain and link it in the final version of this article. The link was removed in the review phase to preserve anonymity. 

\paragraph{Baseline}
We compared our algorithm's runtime with the treewidth-based parameterized approach of~\cite{bankselection}, which is the current 
state-of-the-art in bank selection. That approach has an asymptotic runtime of $O(|G|\cdot \texttt{tw}(G)\cdot|D|^{\texttt{tw}(G)+1})$ where $\texttt{tw}(G)$ is the treewidth of $G,$ which can theoretically be up to $7$~\cite{treewidth}. 
We did not consider other previous methods since they either have exponential runtime dependence on the program size~\cite{MinBankSel,MinBankSel2} or no guarantee on the quality or optimality of the results~\cite{JointBankSel}.

\paragraph{Machine}
All experiments were performed on a virtual machine with Oracle Linux (ARM 64-bit), equipped with 1 core of an Apple M2 processor and 4GB of RAM.

\paragraph{Benchmarks}
We exactly followed the setup of~\cite{bankselection}, utilizing the SDCC regression test suite as our benchmark set. 
This suite comprises a total of $13,463$ instances for HC08, $179,611$ instances for Z80, and $84,700$ instances for MCS51. 
These benchmarks are embedded programs expected to operate in resource-constrained environments. Therefore, there is a strong focus is
on code size optimization. Thus, we used the following standard cost function:
\begin{itemize}
    \item \( c(e, b, b) = c(e, b, \perp) = 0;\) 
    \item \( c(e, b_0, b_1) = 1 \)  when \( e \) is an edge from a taken conditional branch;
    \item For all other cases, \( c(e, b_0, b_1) = 1 \).
\end{itemize}
We only compare the runtimes. There is no output comparison since both our approach and~\cite{bankselection} find an 
optimal solution for bank selection and the outputs coincide.

\begin{figure}[H]
	\centering
	\subfloat[HC08]{
		\includegraphics[clip,width=.33\textwidth]{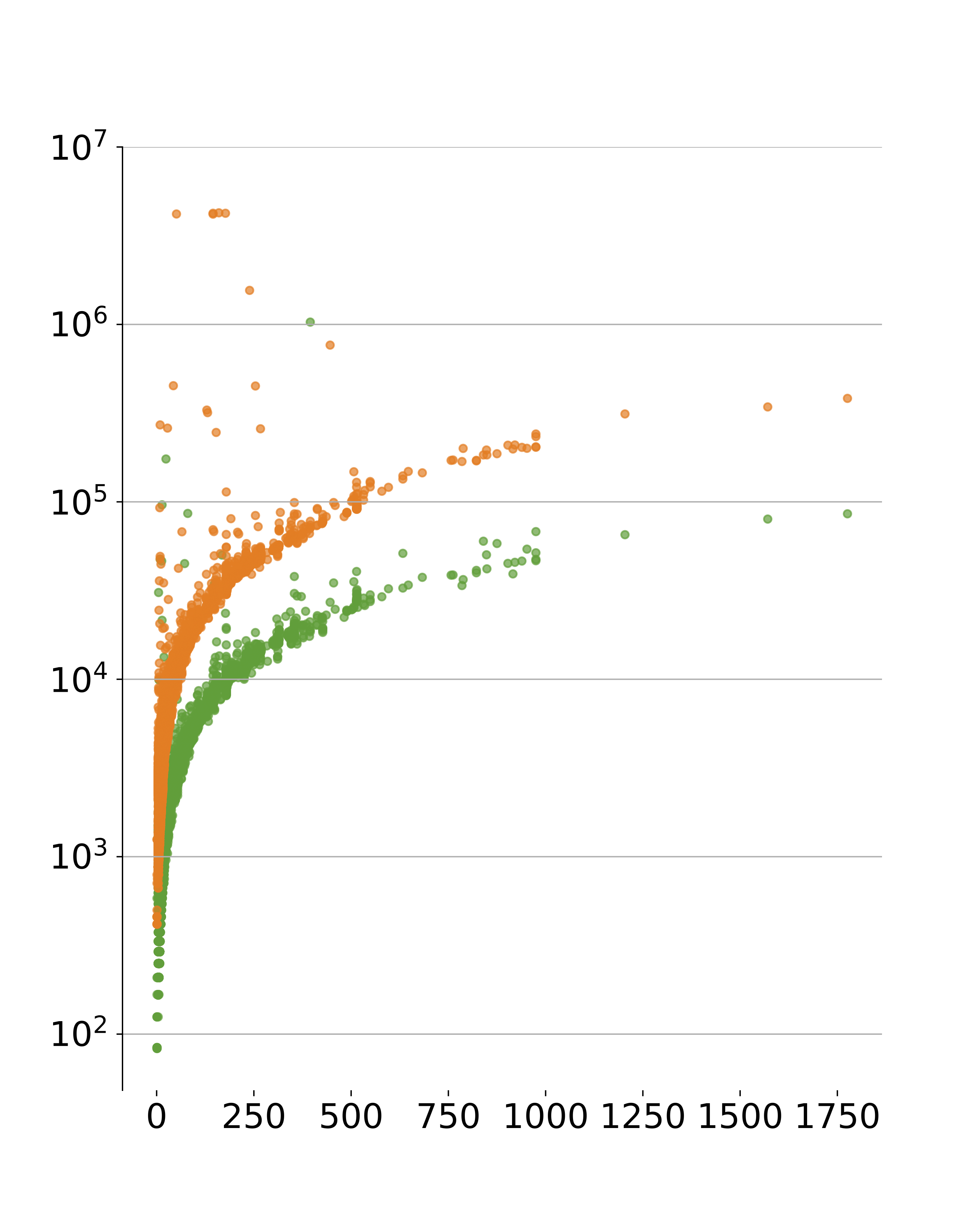}
	}%
	\subfloat[Z80]{
		\includegraphics[clip,width=.33\textwidth]{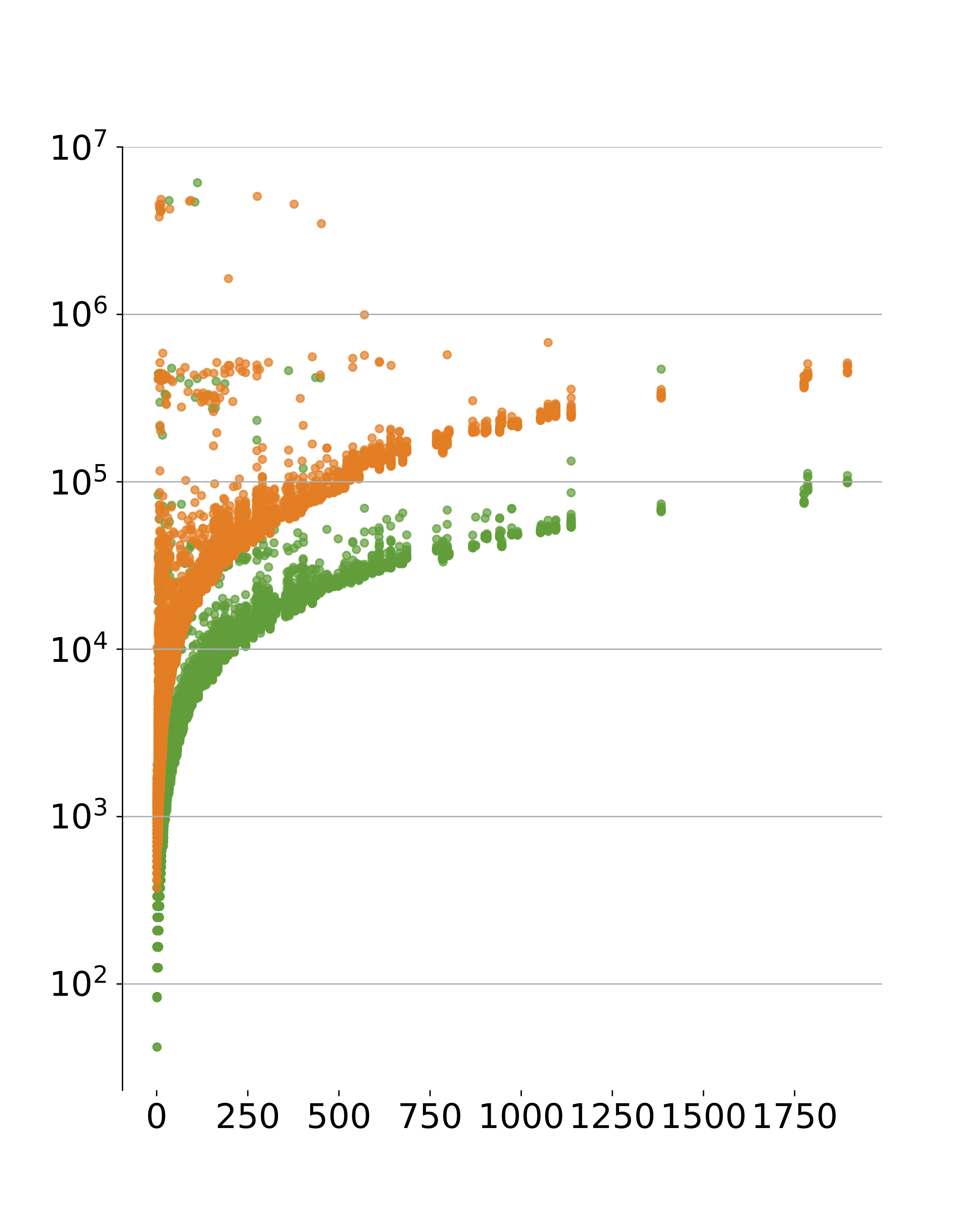}
	}%
	\subfloat[MCS51]{
		\includegraphics[clip,width=.33\textwidth]{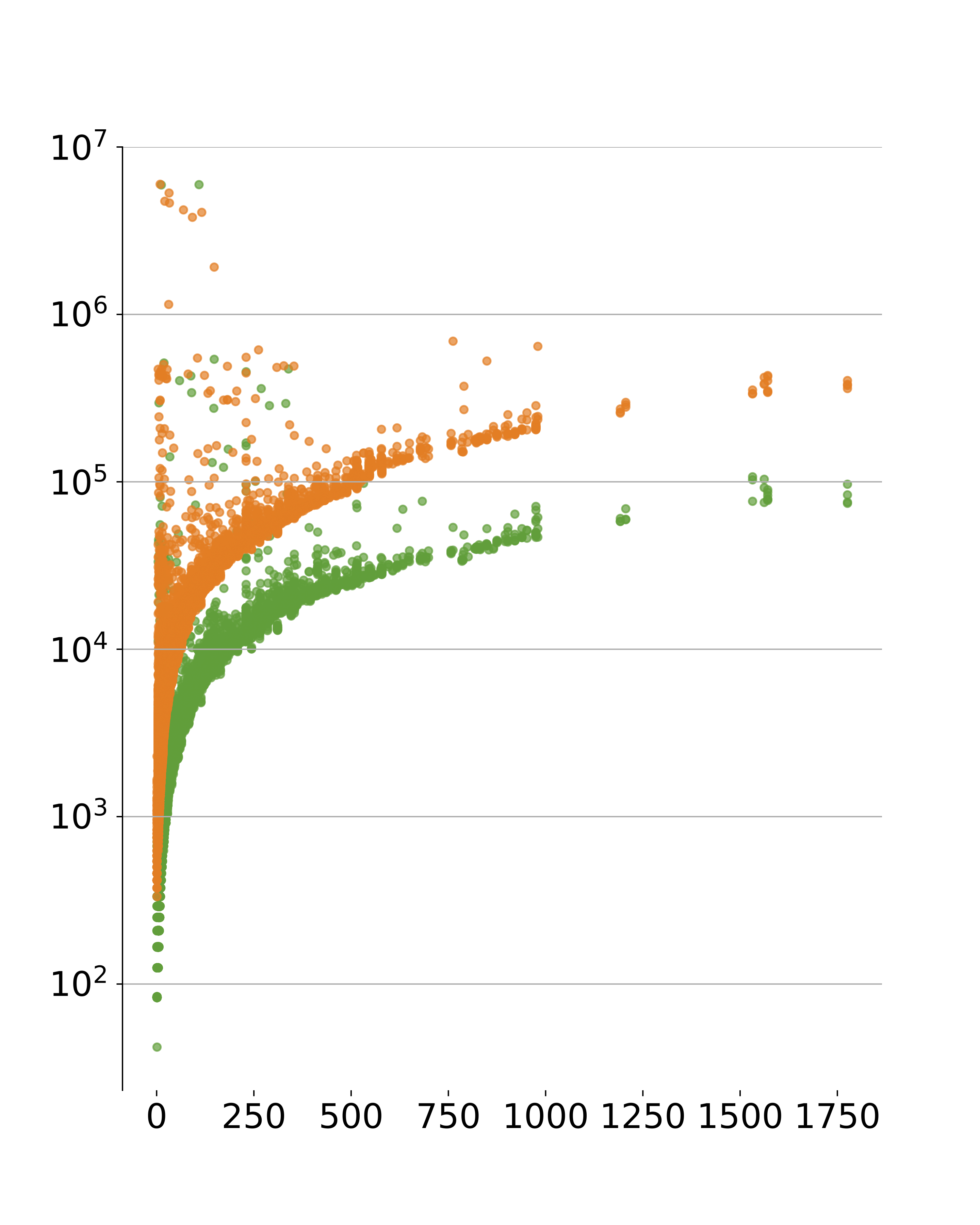}
	}
	\caption{Runtime comparison of the treewidth-based algorithm of~\cite{bankselection} (orange) and our approach (green). The $x$ axis is the number of vertices in the CFG and the $y$ axis is the time in nanoseconds. The $y$ axis is in logarithmic scale.}
	\label{fig:runtime-histogram}
\end{figure}

\begin{figure}[H]
\centering
		\includegraphics[clip,width=\textwidth]{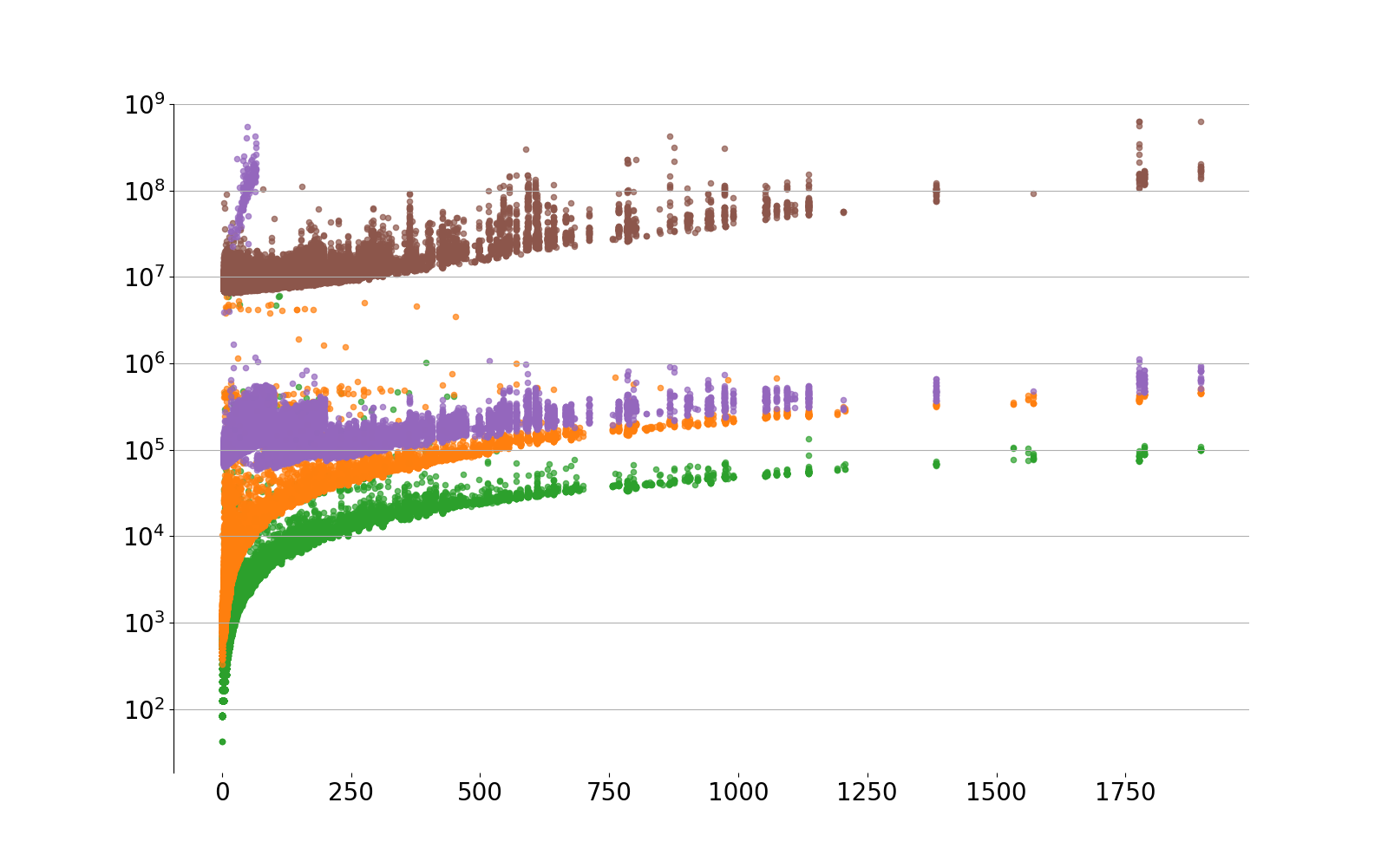}
	\caption{Runtime comparison of the treewidth-based algorithm of~\cite{bankselection} (orange) and our approach (green) compared to SAT-based (brown) and ILP-based (purple) methods.The $x$ axis is the number of vertices in the CFG and the $y$ axis is the time in nanoseconds. The $y$ axis is in logarithmic scale.}
	\label{fig:satilp}
\end{figure}

\paragraph{Runtimes}
Figure~\ref{fig:runtime-histogram} provides a comparison of the runtimes of our algorithm and that of~\cite{bankselection} over the three different architectures. For HC08, our algorithm took an average of $ 1,998.3$ nanoseconds, while the treewidth-based approach took an 
average of $8,558.8$ nanoseconds. For Z80, our algorithm took an average of $ 2,402.1$ nanoseconds, while the 
treewidth-based approach took an average of $8,209.3$ nanoseconds. For MCS51, our algorithm took an average of
$1,834.3$ nanoseconds, while the treewidth-based approach took an average of $6,268.3$ nanoseconds.
\paragraph{Comparison with SAT and ILP Solvers}
Since Optimal Bank Selection is NP-complete, we compare our runtimes with state-of-the-art approaches for NP-complete problems via SAT and ILP solving. We encoded the problem in the natural way in both SAT and ILP. We then used the state-of-the-art SAT solver, Kissat~\cite{sat}, and the industrial ILP solver Gurobi~\cite{gurobi} over the benchmarks. The results, as shown in Figure~\ref{fig:satilp}, illustrate that our method is nearly 10 times faster than the ILP approach, which is itself approximately 100 times faster than the SAT solver.
\paragraph{Discussion}
In summary, our approach is approximately four times faster than the previous state-of-the-art for bank selection. 
This is a notable improvement, particularly given that the treewidth-based  approach of~\cite{bankselection} 
is already highly optimized and included in the well-established SDCC compiler. We believe the speed-up is mainly because our algorithm's runtime does not depend on parameters such as treewidth and the constant factor hidden in our $O(n)$ asymptotic runtime analysis is small in practice.

%% file: sections/conclusion.tex
\section{Conclusion}

In this work, we present an efficient linear-time algorithm for graph PCSPs over control-flow graphs, leveraging SPL decompositions~\cite{RegisterAllocation}. Our solution applies to various compiler optimization tasks, including register allocation~\cite{RegisterAllocation}, Lifetime-optimal Speculative Partial Redundancy Elimination (LOSPRE)~\cite{cai2024faster}, and Optimal Bank Selection. The main contribution is a dynamic programming algorithm that addresses all graph PCSP problems, eliminating the need for separate algorithms for different optimization tasks. Our solution demonstrates significant theoretical performance improvements for bank selection, and experiments show it is approximately four times faster than the treewidth-based approach of~\cite{bankselection} in practice.

%% file: sections/refs.bib
@article{PCS,
	author = {Freuder, Eugene C. and Wallacex, Richard J.},
	title = {Partial Constraint Satisfaction},
	year = {1992},
	booktitle = {{AIJ}},
    volume       = {58},
    pages        = {21-70},
    doi={https://doi.org/10.1016/0004-3702(92)90004-H},
      publisher={Elsevier}

}

@inproceding{bankselection,
    author={Krause, Philipp K.},
    title={Optimal Placement of Bank Selection Instructions in Polynomial Time},
    year={2013},
    booktitle={{M-SCOPES}},
    doi={https://doi.org/10.1145/2463596.2463598},

}

@article{treewidth,
author = {Thorup, Mikkel},
title = {All structured programs have small tree width and good register allocation},
year = {1998},
volume = {142},
doi = {https://doi.org/10.1006/inco.1997.2697},
journal = {Inf. Comput.},
pages = {159–181},
publisher={Elsevier}

}

@misc{sdcc2,
	title={Small Device {C} Compiler},
	author = {Sandeep Dutta and Daniel Drotos and Kevin Vigor and others},
	year={2003},
	url={http://sdcc.sourceforge.net/}
}

@article{sdcc1,
  title={Anatomy of a Compiler: A Retargetable ANSI-C Compiler},
  author={Dutta, Sandeep},
  journal={Circuit Cellar},
  volume={121},
  number={5},
  year={2000}
}

@inproceedings{MinBankSel,
author = {Scholz, Bernhard and Burgstaller, Bernd and Xue, Jingling},
title = {Minimizing bank selection instructions for partitioned memory architecture},
year = {2006},
doi = {https://doi.org/10.1145/1176760.1176786},
booktitle = {{CASES}},
pages = {201–211},
publisher={{ACM}},
}

@article{MinBankSel2,
author = {Scholz, Bernhard and Burgstaller, Bernd and Xue, Jingling},
title = {Minimal placement of bank selection instructions for partitioned memory architectures},
year = {2008},
volume = {7},
doi = {https://doi.org/10.1145/1331331.1331336},
journal = {ACM Trans. Embed. Comput. Syst.},
publisher={{ACM}},
}

@article{JointBankSel,
author = {Liu, Tiantian and Xue, Chun Jason and Li, Minming},
title = {Joint variable partitioning and bank selection instruction optimization for partitioned memory architectures},
year = {2013},
volume = {12},
doi = {https://doi.org/10.1145/2442116.2442126},
journal = {ACM Trans. Embed. Comput. Syst.},
publisher={{ACM}},
}

@inproceedings{cai2024faster,
  title={Faster lifetime-optimal speculative partial redundancy elimination for goto-free programs},
  author={Cai, Xuran and Goharshady, Amir},
  booktitle={{SETTA}},
  pages={382--398},
  year={2024},
  publisher = {Springer},
}

@inproceedings{RegisterAllocation,
author = {Cai, Xuran and Goharshady, Amir Kafshdar and Hitarth, S. and Lam, Chun Kit},
title = {Faster Chaitin-like Register Allocation via Grammatical Decompositions of Control-Flow Graphs},
year = {2025},
doi = {https://doi.org/10.1145/3669940.3707286},
booktitle = {{ASPLOS}},
pages = {463–477},
publisher = {{ACM}},
}

@inproceedings{DBLP:conf/alenex/GustedtMT02,
  author       = {Jens Gustedt and
                  Ole A. M{\ae}hle and
                  Jan Arne Telle},
  title        = {The Treewidth of Java Programs},
  booktitle    = {{ALENEX}},
  series       = {Lecture Notes in Computer Science},
  volume       = {2409},
  pages        = {86--97},
  publisher    = {Springer},
  year         = {2002}
}

@article{Koster2002SolvingCS,
  title={Solving Partial Constraint Satisfaction problems with tree-decomposition},
  author={Arie M. C. A. Koster and Stan P. M. van Hoesel and Antoon W. J. Kolen},
  year={2002},
  doi={https://doi.org/10.1002/net.10046},
  booktitle={Networks},
  volume = {40},
  pages={170-180}
}

@article{KOSTER199889,
title = {The partial constraint satisfaction problem: Facets and lifting theorems},
journal = {Operations Research Letters},
volume = {23},
number = {3},
pages = {89-97},
year = {1998},
doi = {https://doi.org/10.1016/S0167-6377(98)00043-1},
author = {Arie M.C.A. Koster and Stan P.M.van Hoesel and Antoon W.J. Kolen},
}

@article{DECHTER19871,
title = {Network-based heuristics for constraint-satisfaction problems},
journal = {Artificial Intelligence},
volume = {34},
number = {1},
pages = {1-38},
year = {1987},
doi = {https://doi.org/10.1016/0004-3702(87)90002-6},
author = {Rina Dechter and Judea Pearl},
}

@inproceedings{freuder1990complexity,
  title={Complexity of K-tree structured constraint satisfaction problems},
  author={Freuder, Eugene C},
  booktitle={{AAAI}},
  pages={4--9},
  year={1990}
}

@article{MACKWORTH198565,
title = {The complexity of some polynomial network consistency algorithms for constraint satisfaction problems},
journal = {Artificial Intelligence},
volume = {25},
pages = {65-74},
year = {1985},
doi = {https://doi.org/10.1016/0004-3702(85)90041-4},
author = {Alan K. Mackworth and Eugene C. Freuder},
}

@inproceedings{10.5555/2887965.2887992,
author = {Dechter, Rina and Pearl, Judea},
title = {Tree-clustering schemes for constraint-processing},
year = {1988},
booktitle = {{AAAI}},
pages = {150–154},
}

@inproceedings{DBLP:conf/cav/Obdrzalek03,
  author       = {Jan Obdrz{\'{a}}lek},
  title        = {Fast Mu-Calculus Model Checking when Tree-Width Is Bounded},
  booktitle    = {{CAV}},
  pages        = {80--92},
  year         = {2003}
}

@inproceedings{DBLP:conf/esop/ChatterjeeGIP20,
  author       = {Krishnendu Chatterjee and
                  Amir Kafshdar Goharshady and
                  Rasmus Ibsen{-}Jensen and
                  Andreas Pavlogiannis},
  title        = {Optimal and Perfectly Parallel Algorithms for On-demand Data-Flow
                  Analysis},
  booktitle    = {{ESOP}},
  pages        = {112--140},
  year         = {2020}
}

@inproceedings{DBLP:conf/vmcai/GoharshadyZ23,
  author       = {Amir Kafshdar Goharshady and
                  Ahmed Khaled Zaher},
  title        = {Efficient Interprocedural Data-Flow Analysis Using Treedepth and Treewidth},
  booktitle    = {{VMCAI}},
  volume       = {13881},
  pages        = {177--202},
  year         = {2023}
}

@inproceedings{DBLP:conf/cav/ChatterjeeL13,
  author       = {Krishnendu Chatterjee and
                  Jakub Lacki},
  title        = {Faster Algorithms for Markov Decision Processes with Low Treewidth},
  booktitle    = {{CAV}},
  pages        = {543--558},
  year         = {2013}
}

@inproceedings{DBLP:conf/atva/AsadiCGMP20,
  author       = {Ali Asadi and
                  Krishnendu Chatterjee and
                  Amir Kafshdar Goharshady and
                  Kiarash Mohammadi and
                  Andreas Pavlogiannis},
  title        = {Faster Algorithms for Quantitative Analysis of {MCs} and {MDPs} with Small
                  Treewidth},
  booktitle    = {{ATVA}},
  volume       = {12302},
  pages        = {253--270},
  year         = {2020}
}

@inproceedings{DBLP:conf/fsttcs/AhmadiCGMSZ22,
  author       = {Ali Ahmadi and
                  Krishnendu Chatterjee and
                  Amir Kafshdar Goharshady and
                  Tobias Meggendorfer and
                  Roodabeh Safavi and
                  {\DH}orde Zikelic},
  title        = {Algorithms and Hardness Results for Computing Cores of Markov Chains},
  booktitle    = {{FSTTCS}},
  pages        = {29:1--29:20},
  year         = {2022}
}

@inproceedings{DBLP:conf/cav/000120,
  author       = {Sriram Sankaranarayanan},
  title        = {Reachability Analysis Using Message Passing over Tree Decompositions},
  booktitle    = {{CAV}},
  pages        = {604--628},
  year         = {2020}
}

@article{DBLP:journals/toplas/ChatterjeeGGIP19,
  author       = {Krishnendu Chatterjee and
                  Amir Kafshdar Goharshady and
                  Prateesh Goyal and
                  Rasmus Ibsen{-}Jensen and
                  Andreas Pavlogiannis},
  title        = {Faster Algorithms for Dynamic Algebraic Queries in Basic RSMs with
                  Constant Treewidth},
  journal      = {{ACM} Trans. Program. Lang. Syst.},
  volume       = {41},
  number       = {4},
  pages        = {23:1--23:46},
  year         = {2019}
}

@article{DBLP:journals/pacmpl/ConradoGKTZ23,
  author       = {Giovanna Kobus Conrado and
                  Amir Kafshdar Goharshady and
                  Kerim Kochekov and
                  Yun Chen Tsai and
                  Ahmed Khaled Zaher},
  title        = {Exploiting the Sparseness of Control-Flow and Call Graphs for Efficient
                  and On-Demand Algebraic Program Analysis},
  journal      = {Proc. {ACM} Program. Lang.},
  volume       = {7},
  number       = {{OOPSLA2}},
  pages        = {1993--2022},
  year         = {2023}
}

@article{DBLP:journals/toplas/ChatterjeeIGP18,
  author       = {Krishnendu Chatterjee and
                  Rasmus Ibsen{-}Jensen and
                  Amir Kafshdar Goharshady and
                  Andreas Pavlogiannis},
  title        = {Algorithms for Algebraic Path Properties in Concurrent Systems of
                  Constant Treewidth Components},
  journal      = {{ACM} Trans. Program. Lang. Syst.},
  volume       = {40},
  number       = {3},
  pages        = {9:1--9:43},
  year         = {2018}
}

@inproceedings{DBLP:conf/popl/ChatterjeeGIP16,
  author       = {Krishnendu Chatterjee and
                  Amir Kafshdar Goharshady and
                  Rasmus Ibsen{-}Jensen and
                  Andreas Pavlogiannis},
  title        = {Algorithms for algebraic path properties in concurrent systems of
                  constant treewidth components},
  booktitle    = {{POPL}},
  pages        = {733--747},
  year         = {2016}
}

@inproceedings{DBLP:conf/cc/Krause13,
  author       = {Philipp Klaus Krause},
  title        = {Optimal Register Allocation in Polynomial Time},
  booktitle    = {{CC}},
  volume       = {7791},
  pages        = {1--20},
  year         = {2013}
}

@inproceedings{DBLP:conf/soda/BodlaenderGT98,
  author       = {Hans L. Bodlaender and
                  Jens Gustedt and
                  Jan Arne Telle},
  title        = {Linear-Time Register Allocation for a Fixed Number of Registers},
  booktitle    = {{SODA}},
  pages        = {574--583},
  year         = {1998}
}

@inproceedings{DBLP:conf/pldi/AhmadiDGP22,
  author       = {Ali Ahmadi and
                  Majid Daliri and
                  Amir Kafshdar Goharshady and
                  Andreas Pavlogiannis},
  title        = {Efficient approximations for cache-conscious data placement},
  booktitle    = {{PLDI}},
  pages        = {857--871},
  year         = {2022}
}

@article{DBLP:journals/pacmpl/ChatterjeeGOP19,
  author       = {Krishnendu Chatterjee and
                  Amir Kafshdar Goharshady and
                  Nastaran Okati and
                  Andreas Pavlogiannis},
  title        = {Efficient parameterized algorithms for data packing},
  journal      = {Proc. {ACM} Program. Lang.},
  volume       = {3},
  number       = {{POPL}},
  pages        = {53:1--53:28},
  year         = {2019}
}

@inproceedings{oopsla24,
	author = {Amir Kafshdar Goharshady and Chun Kit Lam and Lionel Parreaux},
	title = {Fast and Optimal Extraction for Sparse Equality Graphs},
	year = {2024},
	booktitle = {{OOPSLA}}
}

@inproceedings{DBLP:conf/adaEurope/BurgstallerBS04,
  author       = {Bernd Burgstaller and
                  Johann Blieberger and
                  Bernhard Scholz},
  title        = {On the Tree Width of Ada Programs},
  booktitle    = {Ada-Europe},
  pages        = {78--90},
  year         = {2004}
}

@inproceedings{DBLP:conf/atva/ChatterjeeGP17,
  author       = {Krishnendu Chatterjee and
                  Amir Kafshdar Goharshady and
                  Andreas Pavlogiannis},
  title        = {JTDec: {A} Tool for Tree Decompositions in Soot},
  booktitle    = {{ATVA}},
  volume       = {10482},
  pages        = {59--66},
  year         = {2017}
}

@inproceedings{DBLP:conf/sac/ChatterjeeGG19,
  author       = {Krishnendu Chatterjee and
                  Amir Kafshdar Goharshady and
                  Ehsan Kafshdar Goharshady},
  title        = {The treewidth of smart contracts},
  booktitle    = {{SAC}},
  pages        = {400--408},
  publisher    = {{ACM}},
  year         = {2019}
}

@article{DBLP:journals/pacmpl/ConradoGL23,
  author       = {Giovanna Kobus Conrado and
                  Amir Kafshdar Goharshady and
                  Chun Kit Lam},
  title        = {The Bounded Pathwidth of Control-Flow Graphs},
  journal      = {Proc. {ACM} Program. Lang.},
  volume       = {7},
  number       = {{OOPSLA2}},
  pages        = {292--317},
  year         = {2023}
}

@inproceedings{DBLP:conf/lpar/FerraraPV05,
  author       = {Andrea Ferrara and
                  Guoqiang Pan and
                  Moshe Y. Vardi},
  title        = {Treewidth in Verification: Local vs. Global},
  booktitle    = {{LPAR}},
  volume       = {3835},
  pages        = {489--503},
  year         = {2005}
}

@misc{sat,
	title={Kissat-public},
	author = {Institute for Formal Models and Verification},
	year={2025},
	url={https://fmv.jku.at/kissat/}
}

@misc{gurobi,
 author       = {Gurobi Optimization, LLC},
  title        = {Gurobi},
  year         = {2008},
  url          = {https://www.gurobi.com/},
}
